\documentclass{article} %
\usepackage{iclr2027_conference,times}

\usepackage{amsmath,amsfonts,bm}

\def\eqref#1{equation~\ref{#1}}

\def\1{\bm{1}}

\DeclareMathAlphabet{\mathsfit}{\encodingdefault}{\sfdefault}{m}{sl}
\SetMathAlphabet{\mathsfit}{bold}{\encodingdefault}{\sfdefault}{bx}{n}

\usepackage{hyperref}
\usepackage{graphicx}

\usepackage{url}
\usepackage{enumitem} %
\usepackage{booktabs}
\usepackage{cleveref}
\usepackage{amsmath,amssymb}
\usepackage{hyperref}
\usepackage{sidecap}

\title{\centering
Joints, Camera, Action! Co-Generating 3D Motion
and Video for Arbitrary Rigs
\par\vspace{15pt}
}

\author{
\begin{tabular}{@{}c@{}}
\textbf{Noam Rotstein$^{1,2}$ \quad
Yoad Tewel$^{1}$ \quad
Chen Tessler$^{1}$ \quad
Haggai Maron$^{1,2}$} \\
\textbf{Gal Chechik$^{1,3}$ \quad
Or Litany$^{1,2}$ \quad
Yoni Kasten$^{1,2}$ \quad
Yuval Atzmon$^{1}$} \\[6pt]
{\normalfont $^{1}$NVIDIA \quad
$^{2}$Technion -- Israel Institute of Technology}
{\normalfont $^{3}$Bar-Ilan University}
\end{tabular}
}

\colorlet{reddish}{red!70}

\newcommand{\methodname}{\textsc{JCA}}

\newcommand{\edit}[1]{\textcolor{blue}{#1}}

\newcommand{\noam}[1]{}
\newcommand{\orl}[1]{}
\newcommand{\yuval}[1]{}
\newcommand{\claude}[1]{}
\newcommand{\chen}[1]{}
\newcommand{\yoni}[1]{}

\newcommand{\yoad}[1]{}

\newcommand{\oiv}{OiV}

\newcommand{\gal}[1]{}

\newcommand{\galst}[1]{}

\usepackage{xcolor}
\definecolor{darkgreen}{RGB}{0,100,0}
\renewcommand{\paragraph}[1]{\noindent\textbf{#1}}

\iclrfinalcopy %
\begin{document}

\begingroup
\centering
\maketitle
\par
\endgroup

\lhead{}
\renewcommand{\headrulewidth}{0pt}

\begin{figure*}[h]
    \centering
    \vspace{-20pt}
    \includegraphics[width=\textwidth]{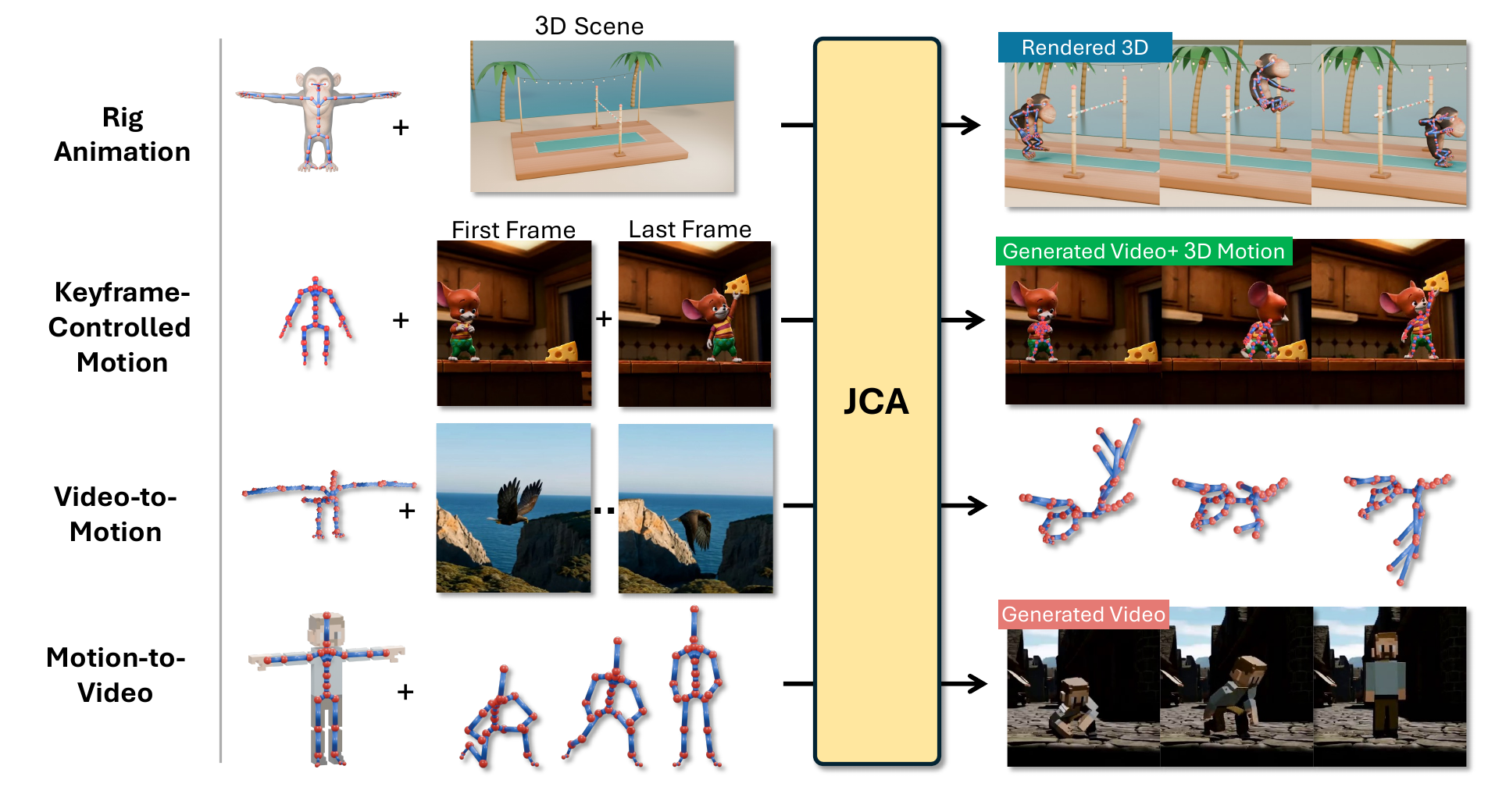}
    \vspace{-26pt}
    \caption{\textbf{Joints, Camera, Action! (JCA)} generates a video and a synchronized 3D motion trajectory for a user-supplied rig of any topology. One model covers joint generation (top two rows) and either conditional direction (bottom two rows).
\textbf{Mesh animation:} a rigged monkey clears a scene’s high-jump bar, adapting its motion to the surroundings.
\textbf{Keyframe control:} given a first and last frame, JCA fills in the rig motion and video between them.
\textbf{Video-to-motion:} recovering 3D motion from an input video.
\textbf{Motion-to-video:} video following prescribed 3D motion.
}
\label{fig:teaser}
\end{figure*}

\begingroup
\setlength{\leftmargini}{3.1em}

\vspace{6pt}
\begin{abstract}
\vspace{-6pt}
Video generation models can now depict remarkably realistic motion.
Yet it is unclear how to use this motion beyond pixels: to animate a specific 3D asset and replay the animation in its 3D scene.
For instance, a video model can show a monkey jumping over a crossbar, but it cannot return that jump as replayable 3D motion placed in the 3D scene.
Here we present a model that directly co-generates the video and 3D motion for any supplied rig, along with a camera that registers this motion to the scene. Prior video--motion models either handle only human skeletons, often adding billions of parameters, or triangulate 3D motion from generated 2D pose maps. We instead keep 3D motion in a separate model that supports arbitrary skeletons, and connect it to a pretrained video model through a small set of cross-attention layers. This separation lets the motion model first learn how skeletons move from motion-only data, which is easier to obtain and augment than paired video. Additionally, our model learns to locate each joint in the video frames, which lets us recover the camera and place the 3D motion in the scene. To train across many skeletons, we use a generative video editor to place animated 3D assets in scenes suited to their actions, curating the edits so their known motion remains valid. Together with human and robot video, this yields nearly 200K training pairs. Our evaluations demonstrate that our model animates diverse rigs from text more faithfully than video-first pipelines, and its motion responds to the surrounding scene. It also turns video into rig motion, rig motion into video, and enables keyframe-based control of the 3D motion.
Videos and additional results are available on our
\href{https://research.nvidia.com/labs/par/projects/jca/}{\edit{project page}}.
\end{abstract}
\endgroup

\section{Introduction}
\label{sec:intro}

Video generation models capture complex real-world dynamics, making them promising world models for animation, robotics, and synthetic data generation. Yet they produce only pixels, so their motion cannot directly drive a 3D asset or robot. Imagine placing a monkey asset in front of a crossbar in a 3D scene, asking a video model to make it jump over, and replaying the resulting jump on the monkey's rig inside that scene. Current video models deliver only the first step: a convincing video of the jump, with no 3D motion for the rig and no way to place that motion in the scene.

Replaying a generated motion on an arbitrary rig in a 3D scene requires two things: 3D motion that matches the video and a camera transformation that registers that motion to the generated scene. One natural approach is to recover the motion from a generated video post hoc~\citep{actionmesh,puppeteer}, but lifting motion from pixels may introduce errors that can make the animation drift from the video. Alternatively, generating video and motion together aims to avoid this post-hoc step, yet no joint model animates an arbitrary rig directly in 3D, and none registers its motion to the scene it generates. Existing ones either handle only human skeletons~\citep{unimo,echomotion,comovi}, or triangulate 3D motion from 2D pose maps learned from multi-view videos~\citep{animax}.

In this work, we present \emph{Joints, Camera, Action!} (\methodname{}), a model that co-generates a video, motion for any supplied 3D rig, and a camera that places the 3D motion in the generated scene. To support arbitrary rigs, a dedicated motion model treats a rig skeleton as a graph, with joints as nodes and bones as edges. We pretrain it on motion alone, including motions retargeted to randomized skeletons. We then connect it to a pretrained video model through a few cross-attention layers, keeping the added parameters small. Zero-initializing these layers preserves the behavior of both pretrained models at the start of coupled training.

Through these layers, video and motion shape each other's generation throughout denoising; two components strengthen this interaction and place the motion in the scene. First, we introduce a novel bridge guidance mechanism. At inference, we contrast predictions made with the connections enabled and disabled, then amplify the change caused by their interaction. %
Co-generating the motion with the video, however, does not determine where that motion lies in the depicted scene. Second, to solve this registration problem, we introduce learned localization of the rig's joints:  %
the model predicts where every generated 3D joint appears in the image. These 2D locations and the generated 3D positions provide correspondences from which we fit a camera that registers the motion to the generated scene.

Training the model poses a challenge: no corpus pairs natural video with 3D motion across many rigs. We start from human video, which offers rich motion and rich backgrounds, but only a single skeleton. To cover many rigs, we turn to animated 3D assets~\citep{deitke2023objaverse, deitke2023objaversexl}, which offer diverse rigs and precise labels, but simple actions and plain renders. Rather than build a 3D scene for every asset, a generative video editor~\citep{bernini} places our animated-asset renders in scenes suited to their actions. Editing may move the subject or camera, invalidating 3D labels; we verify results and regenerate failures. Together, the two sources combine the rich motion and scenes of human video with the rig diversity of assets. With additional robot video, our corpus reaches 200K video--motion pairs in total.

We evaluate the model on text-driven animation of diverse rigs and demonstrate motion that responds to the surrounding scene (\cref{fig:teaser}). The same model can recover motion for a supplied rig from video or generate video from prescribed rig motion. From a few reference image keyframes, it generates a complete 3D rig trajectory and matching video, providing image-based control over 3D motion.

Our contributions are: (1) \methodname{}, the first model to directly co-generate video and 3D joint motion for arbitrary rigs, with motion that responds to the surrounding scene and can be controlled by image keyframes; the same model also generates either modality from the other; %
 (2) a parameter-efficient architecture that represents arbitrary rigs as graphs, with bridge guidance amplifying video--motion agreement; (3) learned scene registration that recovers the camera needed to replay generated 3D motion in the scene; and (4) a data pipeline that places labeled asset animations in scenes suited to their actions and, together with human and robot video, yields  200K video--motion pairs spanning diverse rigs.

\section{Related Work}
\label{sec:related_work}

\textbf{Joint Video and Motion Generation.}
Video generators capture rich appearance and motion and have shown promise across diverse tasks~\citep{rotstein2025pathways, wiedemer2025video, atzmon2025identity}.
However, they output pixels rather than the underlying 3D motion.
UniMo~\citep{unimo}, EchoMotion~\citep{echomotion}, and CoMoVi~\citep{comovi} jointly model video and human motion; SViMo~\citep{svimo} and VAG~\citep{vag} cover hand--object interactions and robot actions.
UniMo uses a shared autoregressive backbone, whereas EchoMotion and CoMoVi introduce billion-scale motion branches derived from video backbones.
EchoMotion's joint attention assigns motion tokens spatial offsets relative to the video grid.
AnimaX~\citep{animax} handles arbitrary skeletons by generating RGB and 2D pose maps from four fixed views, then recovers 3D motion through triangulation and inverse kinematics. This indirect route also limits training to multi-view videos, which rich scenes rarely have.
We couple separately pretrained video and text-to-rig-motion networks through cross-attention, avoiding joint self-attention and spatial offsets.
Our model directly generates video and 3D joint parameters for a supplied rig, without mesh input or multi-view triangulation, optionally conditioned on a first image.

\noindent
\textbf{Animating Diverse Skeletons and Meshes.}
AnyTop~\citep{anytop} learns a shared motion prior across diverse skeletons; RigMo~\citep{rigmo} learns rigs and motion from deforming meshes without predefined skeletons. AnimateAnyMesh~\citep{animateanymesh} maps mesh and text to vertex trajectories.
Video priors inform ViPS's feed-forward pose sampling~\citep{vips} and per-instance animation optimization by AKD~\citep{akd}, Puppeteer~\citep{puppeteer}, and skeleton-free CHORD~\citep{choreographing}; AKD reports about 25 hours per asset. AnimaSpark~\citep{animaspark} avoids such optimization through video generation, tracking, and planar lifting. ActionMesh~\citep{actionmesh} and STAC ~\citep{stac} output mesh deformations and~\cite{gsgd} transfers motion across object categories without a rig by distilling video diffusion priors into dynamic Gaussians.
Our model jointly generates motion and video through coupled denoising, without tracking or fitting the motion to video.

\noindent
\textbf{Conditional Generation and Guidance.}
Structural and motion cues can guide visual generation:
ControlNet~\citep{controlnet} conditions image synthesis on structural input signals, Lay-A-Scene~\citep{layascene} on 3D objects, MeshToken~\citep{meshtoken} conditions video on 3D human mesh motion, LooseControlVideo~\citep{loosecontrolvideo} uses coarse oriented 3D boxes, and training-free Time-to-Move~\citep{ttm} uses crude reference animations under a dual-clock denoising schedule.
Motion inpainting and constrained generation~\citep{maskedmimic, kimodo} complete partial motion specifications.
EchoMotion~\citep{echomotion} supports video-to-motion and motion-to-video generation for humans.
Our training combines joint denoising with conditioning on a noise-free modality, enabling both conditional directions for arbitrary rigs through our specialized rig encoder and denoiser.
To improve generation at inference, latent optimization corrects spatial relations~\citep{learntosteer}, VideoJAM~\citep{videojam} contrasts predictions with and without the jointly generated optical-flow input, and Spatiotemporal Skip Guidance~\citep{stg} contrasts predictions with and without selected spatiotemporal layers.
Our bridge guidance strengthens video--motion consistency by contrasting predictions with video--motion cross-attention enabled and disabled.

\noindent
\textbf{Motion Recovery.}
While human-specific methods recover body~\citep{sambody4d,gvhmr} and hand motion~\citep{dynhamr} from video, MoCapAnything V2~\citep{mocapanythingv2} and TopoCap~\citep{topocap} support arbitrary skeletons without per-instance optimization.
Our model also supports this video-to-motion task within its joint generative framework.

\noindent
\textbf{Training Data Pipeline.}
Joint video--motion methods~\citep{unimo,echomotion,comovi} train on human videos paired with estimated motion, using a fixed skeletal representation.
VideoMDM~\citep{videomdm} avoids 3D annotation through 2D projection supervision, but also assumes a fixed skeleton.
Animated asset datasets such as AnimaX's~\citep{animax} provide motion for diverse skeletons, but their renders have a white background.
Generative models have enabled the construction of training pairs for image and video editing~\citep{pipe,veggie}, and video editors can change a video's scene while preserving its structure~\citep{bernini,byg}. We use Bernini to place rendered animations of diverse rigs in suitable scenes while retaining their known 3D motion as supervision.

\section{Method}
\label{sec:method}
\vspace{-5pt}

Video models have learned from countless examples of walking, crawling, and leaping, yet lack an explicit representation of the underlying rigs.
Motion models, in contrast, are built around a rig but do not directly benefit from the rich motion priors learned from video.
We bring the two together: our method generates 3D motion for an arbitrary skeleton by denoising it jointly with a video, so that each model contributes what it knows.

Given a text prompt $c$, a rig defined as a directed tree $\mathcal{G}=(\mathcal{J},\mathcal{E})$, and optionally a reference image $I_0$ with the 2D locations of the rig's joints, we produce a video, a 3D motion for $\mathcal{G}$ given as per-joint rotations and positions, and a camera that projects this motion into the scene that the video depicts.

\begin{figure}[t]
  \centering
  \includegraphics[width=1.\linewidth]{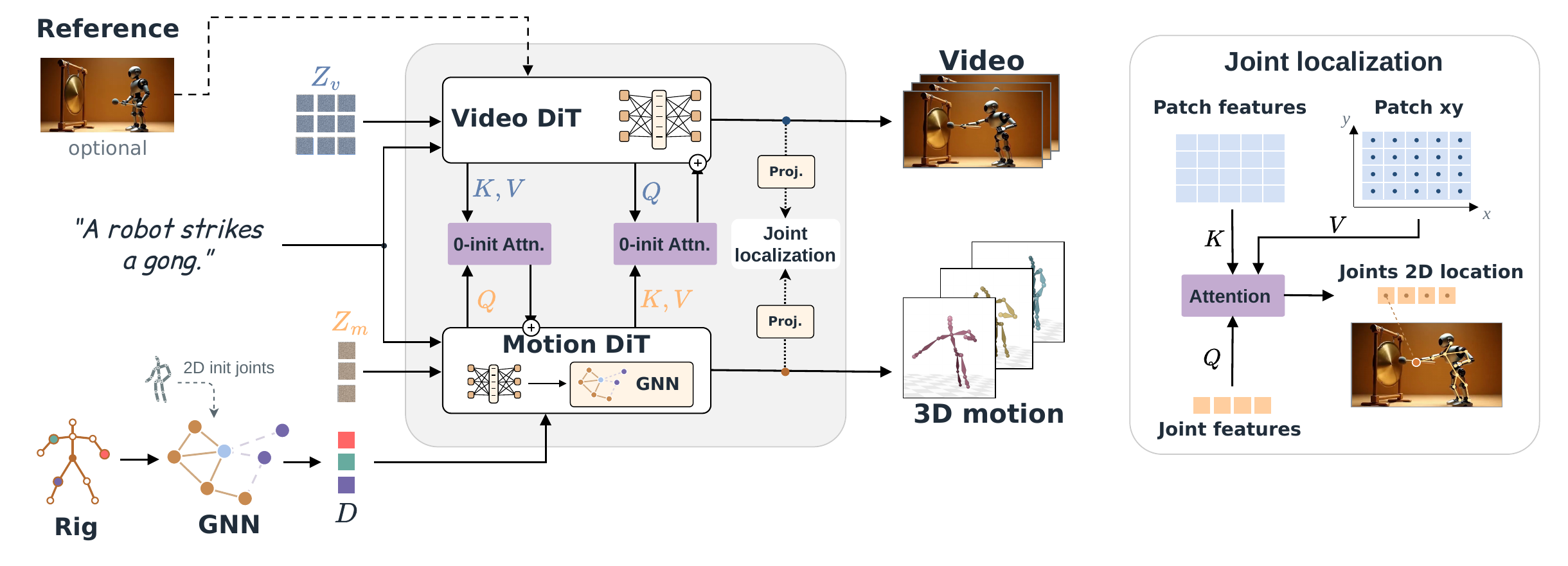}
  \vspace{-16pt}
  \caption{
  Architecture overview.
  Given a text prompt, a rig, and an optional reference image, the model jointly denoises video latents $Z_v$ and motion tokens $Z_m$. A graph network encodes the rig into per-joint descriptors $D$, using the joints' 2D locations in the reference image, and these condition the motion branch; a graph-attention layer inside each Motion DiT block propagates features along the rig edges. Zero-initialized cross-attention bridges exchange features in both directions at paired blocks, each branch querying the other's intermediate features, so joint training begins from the two pretrained branches acting as they did in isolation. An auxiliary joint-localization head (right) attends from joint features to video patches and returns each joint's image position as a weighted combination of patch coordinates, giving  2D--3D correspondences used to fit a camera.}
  \vspace{-8pt}
  \label{fig:architecture}
\end{figure}

We describe our approach in four parts, illustrated in Fig.~\ref{fig:architecture}. 
Sec.~\ref{sec:motion} describes a motion denoiser designed and pretrained for arbitrary rigs, conditioned on per-joint descriptors from a graph-based rig encoder.
Sec.~\ref{sec:crossattn} connects it to a pretrained video generator through zero-initialized bidirectional cross-attention.
This coupling preserves separate backbones, saving billions of parameters compared with motion branches derived from the video backbone, and enables guidance that strengthens their interaction at inference.
Sec.~\ref{sec:localization} introduces a localization head that grounds each joint in the video, from which we fit the camera. Finally, Sec.~\ref{sec:data} describes how we build training data pairing videos with 3D motion across many skeletons, by editing renders of animated assets into suited scenes.

\vspace{-3pt}
\subsection{Rig-Conditioned Motion Generation}
\label{sec:motion}
\vspace{-3pt}

For motion generation, we use a text-conditioned flow-matching network~\citep{lipman2023flow} that generates joint trajectories for rigs with varying joint counts and skeletal connectivity.
A graph encoder maps the rig to per-joint descriptors, which condition a transformer denoiser throughout generation.
We train the rig encoder and motion denoiser from scratch on motion data before coupling them to video.

We define the flow-matching path from a clean sample $z_0$ as $z_s=(1-s)z_0+s\epsilon$, $\epsilon\sim\mathcal{N}(0,I)$, with noise level $s\in[0,1]$ and velocity target $v^{\star}=\epsilon-z_0$.

\paragraph{Motion representation.}
Inspired by AnyTop~\citep{anytop}, we use a redundant representation. Each joint stores a 6D rotation~\citep{zhou2019continuity} and a 3D position, together with their frame-to-frame changes, totaling 18 channels.
A motion with $N$ frames and $J$ joints is represented as $Z_m\in\mathbb{R}^{N\times J\times 18}$.
Root rotations and positions are global; non-root rotations are parent-local, while positions are expressed in the root's coordinate frame. Details are in App.~\ref{app:motion_repr}.

\paragraph{Rig Descriptors.}
Across arbitrary rigs, joint counts and connectivity vary, and joint indices do not carry consistent semantic meaning.
We therefore encode a rig with $J$ joints into descriptors $D\in\mathbb{R}^{J\times d}$, one per joint. Each joint's input combines its name, embedded with the video model's frozen text encoder, its rest-pose geometry, its reference-image location when available, and a root, chain, or leaf indicator. Separate projections embed the text and geometric features, which are concatenated and fused through an MLP.
A graph attention network~\citep{brody2022attentive} propagates the resulting tokens bidirectionally along the bones $\mathcal{E}$, capturing local skeletal relationships. Full self-attention then enables direct
communication between joints that are distant in the hierarchy. The resulting descriptors are shared across frames and added to the corresponding motion tokens through a learned projection at each denoiser block. Since $D$ depends only on the rig and reference-image joint locations, it is computed once and reused throughout sampling (App.~\ref{app:rig_descriptor}).

\paragraph{Motion Denoiser.}
The motion denoiser operates on one token per joint per frame, and predicts the flow velocity, $\hat v = \Phi_\theta(Z_{s}, s, D, \mathcal{E}, c)$, where $Z_s$ denotes the motion $Z_m$ corrupted at noise level $s$.
Each transformer block combines global self-attention over all $NJ$ tokens with graph attention along $\mathcal{E}$ within each frame. Global attention captures interactions across joints and time, while graph attention explicitly incorporates the rig's skeletal connectivity. Text conditions each block through cross-attention, and the noise level $s$ controls adaptive normalization. A block-specific projection of each joint's rig descriptor is added to its tokens before these updates.
Temporal rotary embeddings encode frame indices, with all joints in a frame sharing the same temporal coordinate.
Input and output MLPs map between motion features and hidden tokens, using separate networks for the root and networks shared by all non-root joints.
 Further architectural details are provided in App.~\ref{app:denoiser}.

\vspace{-3pt}
\subsection{Coupling Video and Motion}
\label{sec:crossattn}
\vspace{-3pt}

How should the two models exchange information? 
Prior work on joint video--motion generation extends the video transformer's blocks with motion-specific weights and joint attention over video and motion tokens~\citep{echomotion}.
Applying EchoMotion's hybrid design to half of our video backbone's 40 blocks would add approximately 7B parameters in motion-specific transformer blocks alone.
Our motion denoiser, rig encoder, localization head, and bridges together add only 178M parameters.
We therefore keep the two models separate, and connect the motion denoiser to the pretrained Wan2.2 I2V-A14B video backbone~\citep{wan2025} by inserting cross-attention ``bridges'' in both directions between each motion block and one video block, spread approximately evenly through the backbone. At such a block, let $\bar H_v$ and $\bar H_m$ be the features entering the self-attention of the video and motion branches. We compute
\[
B_{m\rightarrow v}
= \operatorname{Attn}(\bar H_v,\bar H_m)\,O_v,
\qquad
B_{v\rightarrow m}
= \operatorname{Attn}(\bar H_m,\bar H_v)\,O_m \quad,
\]
where $\operatorname{Attn}(X,Z) = \operatorname{softmax}(QK^{\top}/\sqrt{d_k})V$, with queries derived from $X$ and keys and values from $Z$ through direction-specific learned projections.
Each bridge output is added to the receiving branch's self-attention output before its gated residual update.

The output projections $O_v$ and $O_m$  of these bridges are initialized to zero. Joint training thus starts from the two pretrained branches acting  as they did in isolation, and  coupling is learned from there (App.~\ref{app:coupling}).
Following EchoMotion~\citep{echomotion}, we express both streams on a common temporal axis for cross-modal attention, assigning video latent frame $f$ the coordinate $f$ and motion frame $n$ the coordinate $n/4$.

\label{bridge}
\paragraph{Bridge guidance.}
At inference, we retain standard text guidance while adding separate control over cross-modal coupling. Text guidance strengthens prompt adherence, while our bridge guidance amplifies the effect of video--motion interaction.
To achieve this, we compute velocity predictions under two configurations of the same model: with its bidirectional cross-attention bridges disabled and with them enabled.
We apply text guidance to each configuration and then combine the two guided predictions using the bridge guidance scale.
Let $v_i^{\,b}(c)$ denote the velocity predicted for modality $i\in\{\mathrm{video},\mathrm{motion}\}$ under prompt $c$, with bridges disabled ($b=0$) or enabled ($b=1$).
For each bridge configuration $b$, we first apply text guidance using the prompt $c$ and the empty prompt $\varnothing$, $\tilde v_i^{\,b}=v_i^{\,b}(\varnothing) +\omega_{\mathrm{text},i}\big(v_i^{\,b}(c)-v_i^{\,b}(\varnothing)\big)$.
We then scale the change from the disconnected prediction to the connected one, $\hat v_i=\tilde v_i^{\,0}+\omega_{\mathrm{bridge},i}\big(\tilde v_i^{\,1} -\tilde v_i^{\,0}\big)$.
The scale $\omega_{\mathrm{bridge},i}$ sets how strongly modality $i$ follows the other.
Guided sampling requires four forward passes per denoising step.

\subsection{Joint Localization and Camera Alignment}
\label{sec:localization}

The attention bridges discussed above bind the video and motion streams at generation, but do not explicitly locate the generated 3D joints in the video.
We therefore train an auxiliary joint-localization head that predicts their 2D image locations from the final features of both branches.

Each joint at motion frame $4f$ attends only to the patches of video latent frame $f$.
With queries $Q_{4f}$ computed from the joints' motion features at frame $4f$ and keys $K_f$ from the video latent frame's patch features, the weights
$A_f=\operatorname{softmax}(\kappa\,Q_{4f}K_f^{\top})$, with a learned scale $\kappa$, average the patch centers $C$ into image locations $\widehat Y_f = A_f\,C$, supervised by smooth-$L_1$ against 2D annotations.
Pairing $\widehat Y_f$ with the generated 3D joint positions, we fit one static perspective camera per clip, estimating rotation, translation, and focal length while fixing the principal point at the image center.
Further details on the localization head and camera fitting are provided in App.~\ref{app:localization}.
Figure~\ref{fig:scene_cam_align} shows the same generated motion rendered in an input scene before and after alignment, highlighting its importance for correctly positioning the character.

\begin{figure*}[t]
    \centering
    \includegraphics[width=.94\textwidth]{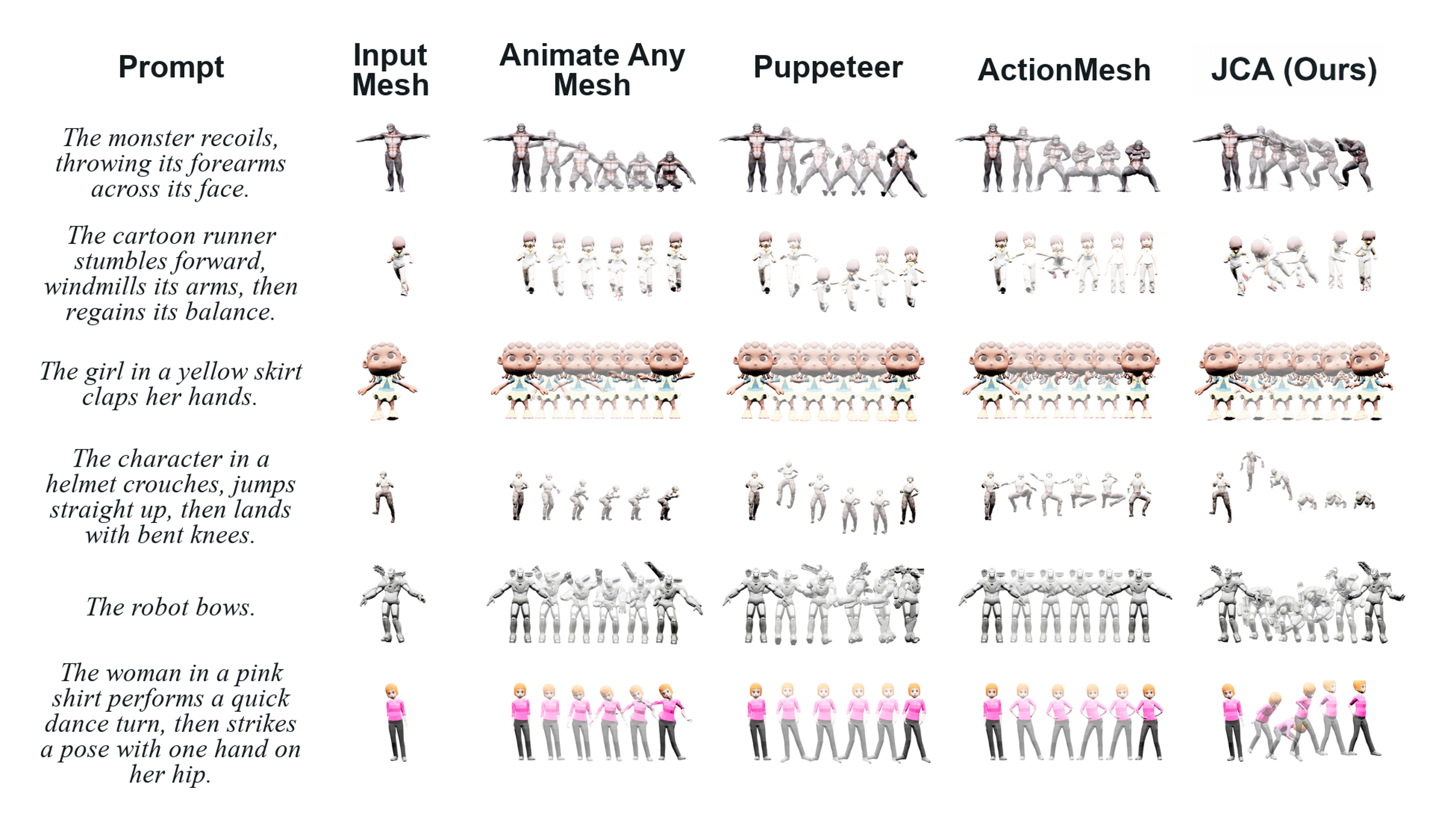}
    \setlength{\abovecaptionskip}{3pt}
    \setlength{\belowcaptionskip}{0pt}
    \vspace{-10pt}
    \caption{Text-driven rig animation across diverse assets. Each row shows a text prompt, an input mesh, and sampled poses from animations generated by competing methods and JCA (ours). Our method more faithfully captures the requested actions while preserving the asset’s geometry.}
    \label{fig:t2m_comp}
    \vspace{-12pt}
\end{figure*}

\vspace{-4pt}
\subsection{Training Data}
\label{sec:data}
\vspace{-4pt}

To our knowledge, there exists no corpus that pairs natural video with 3D motion for arbitrary rigs.
We therefore build and annotate about 200K video--motion pairs from complementary sources.
Human video has rich scenes and interactions, but limited skeletal diversity.
Animated assets from Objaverse~\citep{deitke2023objaverse} and Objaverse-XL~\citep{deitke2023objaversexl} span a wide range of skeletons with known 3D motion, but their animations often depict simple actions and their renders lack natural scene context.

One could place each asset in a 3D scene and render it.
But suitable scenes are scarce, hard to match to each action, and can still look synthetic.
Instead, we let a video model build the world around the motion.
We render each asset in a simple temporary setting, then use a vision-language model to describe its action and propose an environment in which it belongs.
A generative video editing model~\citep{bernini} brings this environment into the rendered clip, pairing the visual richness of a generated scene with the known 3D motion of the source animation.
This turns isolated animated assets into a scalable source of realistic video--motion pairs across diverse rigs.

The central challenge is preserving the source motion and its agreement with the edited video, since editing can change the subject's appearance, pose, or trajectory and invalidate its 3D annotations.
We compare pre- and post-edit masks to detect changes in the subject's silhouette and image-space motion, and use a vision-language model to reject physically implausible results.
We also check that the camera remains static to avoid global image motion caused by camera drift.
Further details on these checks are provided in App.~\ref{app:data}.
Eligible failed clips are regenerated with new random seeds.
By showing assets in realistic scenes, we aim to narrow the visual gap to scenes seen at inference. The resulting dataset, ``Objaverse in Video'' (\oiv{}), contains 118K videos from 43K assets.

We complement \oiv{} with 60K human video clips for which we reconstruct body and hand motion, and 27K robot video clips with motion obtained from their kinematic models, yielding approximately 200K video--motion pairs in total.
For motion-only pretraining, we additionally use 125K motion sequences, including human motions retargeted to randomized humanoid rigs with varying joint counts, connectivity, and geometry.
This leverages the richness of human motion to expand skeletal diversity without requiring additional motion capture or paired videos.
Qualitative examples are available on our \href{https://research.nvidia.com/labs/par/projects/jca/}{project page}.

\vspace{-4pt}
\subsection{Training}
\vspace{-4pt}
\label{sec:training_and_inference}
We train the motion denoiser and rig encoder from scratch, and separately adapt the video backbone with LoRA. Joint training then updates these two, the LoRA adapters, the cross-modal bridges, and the localization head, leaving the base video weights frozen; captions, the reference image with its 2D joint coordinates, and the bridges are randomly dropped, and in some steps one modality is kept clean to serve as a condition. See App.~\ref{app:training_and_inference} for details.

\vspace{-5pt}
\section{Experiments}
\vspace{-4pt}

We demonstrate that \methodname{} outperforms video-first and mesh-based baselines in text-driven rig animation and, after fine-tuning, surpasses a human-specific model for joint motion--video generation. It also enables scene-aware, conditional, keyframe-controlled, and image-based animation. Quantitative evaluations and qualitative comparisons show that its motion follows text faithfully, preserves asset geometry, stays synchronized with generated video, and responds to the surrounding scene.

\vspace{-4pt}
\subsection{Rig Animation}
\label{rig_animation}
\vspace{-4pt}

We evaluate how effectively our generated motion animates diverse rigged assets from text. We randomly sample 20 assets from the Puppeteer test set~\citep{puppeteer} and use GPT-5.6 Sol to generate five action prompts per asset, giving 100 asset--prompt pairs. We compare against AnimateAnyMesh~\citep{animateanymesh} and ActionMesh~\citep{actionmesh}, which animate mesh vertices without a rig, and Puppeteer~\citep{puppeteer}, which animates the ground-truth rig. We exclude AnimaX~\citep{animax} because its code, weights, and evaluation benchmark are not publicly available.
Puppeteer and ActionMesh recover motion from a generated guiding video, representing video-first pipelines. All methods start from the same posed asset:
the baselines receive the 3D asset, whereas
our method receives the rest-pose rig, a rendered image and its 2D joint positions.

\begin{table}[t]
\centering
\small
\setlength{\tabcolsep}{4pt}
\renewcommand{\arraystretch}{0.9}
\begin{minipage}[c]{0.64\textwidth}
\centering
\resizebox{\linewidth}{!}{%
\begin{tabular}{lcccc}
\toprule
\textbf{Method} &
\shortstack{\textbf{Prompt}\\\textbf{adherence} $\uparrow$} &
\shortstack{\textbf{Motion}\\\textbf{smoothness} $\uparrow$} &
\shortstack{\textbf{Dynamic}\\\textbf{degree}} &
\shortstack{\textbf{Mesh}\\\textbf{distortion} $\downarrow$} \\
\midrule
AnimateAnyMesh & 20.6 & \textbf{.993} & .11 & 1.96 \\
Puppeteer      & 41.1 & .984 & .80 & 0.36 \\
ActionMesh     & 43.4 & .988 & .50 & 7.07 \\
\midrule
\methodname{} (ours)           & \textbf{58.4} & .983 & .83 & \textbf{0.20} \\
\bottomrule
\end{tabular}%
}
\end{minipage}\hfill
\begin{minipage}[c]{0.33\textwidth}
\caption{Rig animation from text, averaged over 100 asset--prompt pairs. Prompt adherence is judged by GPT-6 Astra on a 0--100 scale. Mesh distortion is in percent.}
\label{tab:rig_animation}
\end{minipage}
\vspace{-6pt}
\end{table}

\begin{table}[t]
\centering
\setlength{\abovecaptionskip}{3pt}
\setlength{\belowcaptionskip}{2pt}
    \vspace{-4pt}
    \caption{Joint video--human-motion generation. Synchronization compares generated and GVHMR-recovered motion; video and dynamics metrics follow VBench.} %
    \label{tab:echomotion_comparison}
    \setlength{\tabcolsep}{3.5pt}
    \renewcommand{\arraystretch}{0.88}
    \resizebox{\linewidth}{!}{%
    \begin{tabular}{l ccc @{\hspace{10pt}} cccc @{\hspace{10pt}} cc}
    \toprule
    & \multicolumn{3}{c}{\textbf{Synchronization}}
    & \multicolumn{4}{c}{\textbf{Video quality}}
    & \multicolumn{2}{c}{\textbf{Dynamics}} \\
    \cmidrule(lr){2-4} \cmidrule(lr){5-8} \cmidrule(lr){9-10}
    \textbf{Method} &
    \shortstack{PA-\\MPJPE $\downarrow$} &
    \shortstack{SA-\\MPJPE $\downarrow$} &
    \shortstack{SA-\\MPJVE $\downarrow$} &
    \shortstack{\\IQ $\uparrow$} &
    \shortstack{\\Aesth. $\uparrow$} &
    \shortstack{\\Smooth. $\uparrow$} &
    \shortstack{\\Anat. $\uparrow$} &
    \shortstack{Dyn.\\degree} &
    \shortstack{Non-static\\camera} \\
    \midrule
    EchoMotion & 97.7 & 114.9 & 431.3 & \textbf{.665} & .533 & .985 & .730 & .828 & 74.0\% \\
    \midrule
    \methodname{} (ours)       & 114.8 & 151.9 & 598.2 & .664 & \textbf{.578} & .992 & .934 & .600 & 45.6\% \\
    \methodname{} SMPL-FT    & \textbf{72.3} & \textbf{91.0} & \textbf{272.6} & .657 & .573 & \textbf{.994} & \textbf{.938} & .488 & 44.8\% \\
    \bottomrule
    \end{tabular}
    }
    \vspace{-10pt}
\end{table}

\paragraph{Metrics.}
We evaluate prompt adherence with GPT-6 Astra as a multimodal judge, since text-driven motion has no ground truth. The judge scores each animation against its prompt from 0 to 100, blind to method identity (\cref{app:rig_animation}).
We measure how well the asset's geometry is preserved, because  vertex-based methods can violate it, and report mesh distortion: the median absolute change in edge length relative to the first frame. Finally, we report VBench's motion smoothness and dynamic degree~\citep{vbench}, which together separate smooth motion from motion that barely moves.

\paragraph{Results.}
Our method outperforms all baselines in prompt adherence by a wide margin, scoring $58.4$ against $43.4$ for the strongest baseline (\cref{tab:rig_animation,fig:t2m_comp}), with the lowest mesh distortion and the highest dynamic degree. The two strongest baselines recover motion from a video generated without knowledge of the rig, supporting our premise that the rig should inform generation rather than enter only afterward. AnimateAnyMesh is the smoothest, but its low dynamic degree shows that its motion is subtle.

\vspace{-4pt}
\subsection{Scene-Aware Rig Animation}
\label{exp:scene_aware}
\vspace{-4pt}

Prior work on arbitrary-rig animation, as discussed in \cref{sec:related_work}, has focused on isolated meshes without considering their interaction with the surrounding environment.
Our training procedure and video-generation backbone enable our model to generate scene-aware motion.
Given a rigged asset and a 3D scene, we place the asset at a user-specified location and pose and render this configuration as the first frame. From a text prompt, the model then generates a video that respects the scene geometry, together with the corresponding 3D motion. For example, the ape in \cref{fig:teaser} jumps over a crossbar, and the penguin in \cref{fig:scene_cam_align} slides on its belly across the snow (more examples in \href{https://research.nvidia.com/labs/par/projects/jca/}{our page}).

\subsection{Joint Generation for Humans}
\label{exp:human_joint}
\vspace{-6pt}

Our method is designed for arbitrary rigs. When using  a human skeleton, it can be directly compared with EchoMotion~\citep{echomotion}, a human-specific model that jointly generates an RGB video and an SMPL motion sequence from text.
We evaluate two variants: our general model, \methodname{}, and \methodname{} SMPL-FT, obtained by fine-tuning the general checkpoint for 10K steps on our human subset using SMPL, the body model used by EchoMotion.
Both variants are evaluated on 250 prompts from the HumanML3D test split~\citep{humanml3d}.
EchoMotion runs text-to-video, while we run ours with a generated reference image (\cref{app:human_eval}).

\begin{figure*}[t]
    \centering
    \includegraphics[width=\textwidth]{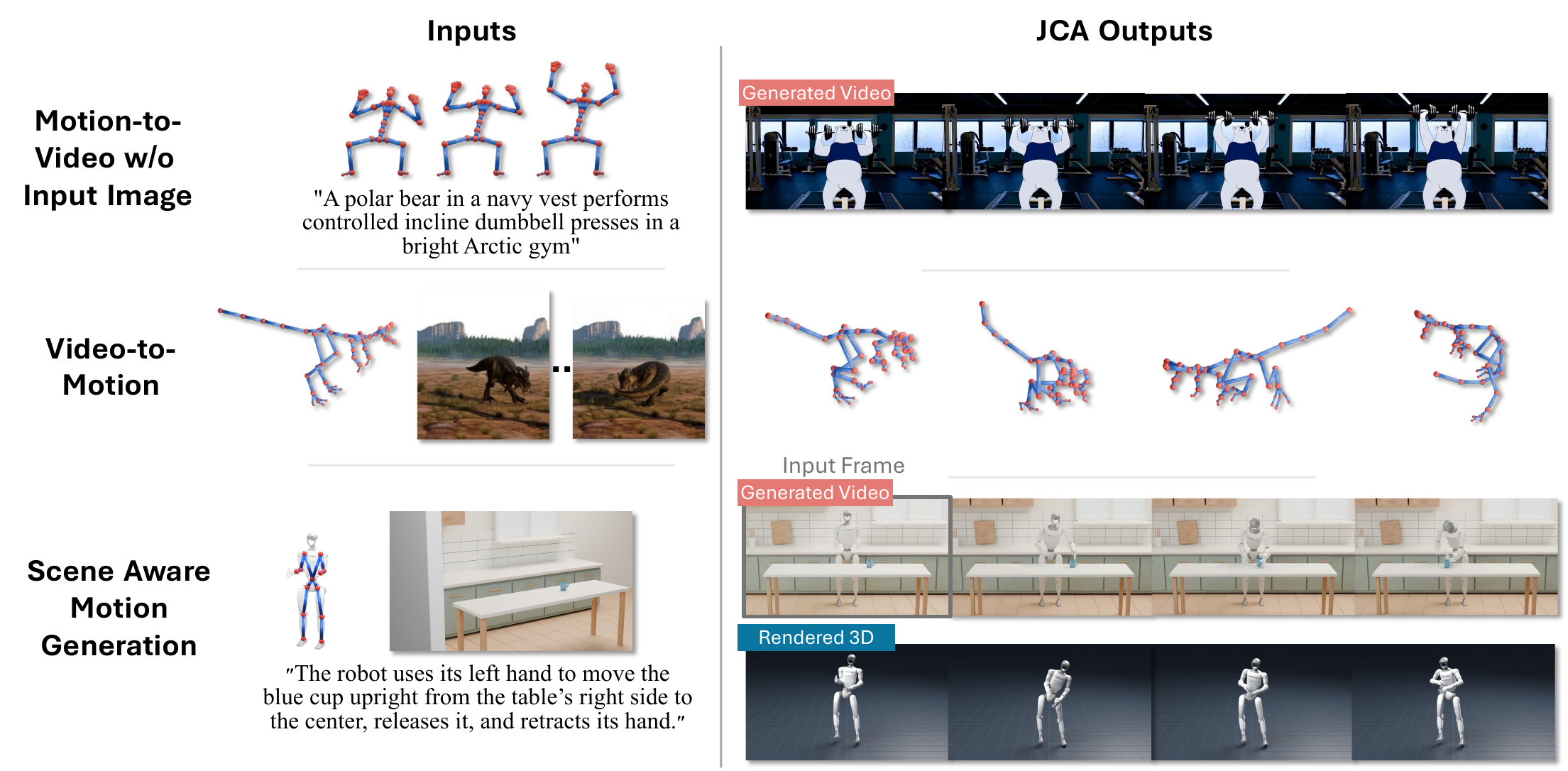}
    \vspace{-18pt}
    \caption{Additional Applications of our model.
    \textbf{Top:} Unlike the setting in \cref{fig:teaser}, we generate a video from a rig motion sequence and a text prompt, without an input image.
    \textbf{Middle:} An additional video-to-motion example with a different skeletal topology.
    \textbf{Bottom:} Given the rig and mesh of a real Unitree G1 robot, a 3D scene, an initial pose, and text, we jointly generate video and motion; we show the video alongside the mesh driven by the generated motion and rendered in the scene.
    }
    \vspace{-10pt}
    \label{fig:extra_results}
\end{figure*}

\paragraph{Metrics.}
We evaluate synchronization, whether the generated motion matches the generated video, by recovering an SMPL trajectory from each video with GVHMR~\citep{gvhmr} and comparing it with the generated one.
PA-MPJPE measures pose agreement after per-frame rigid alignment; SA-MPJPE and SA-MPJVE measure root-centered position and velocity agreement under a single rotation fitted over the sequence.
For video quality, we follow EchoMotion and report VBench metrics, adding the non-static camera rate since dynamic degree also counts camera motion (\cref{app:human_eval}).

\paragraph{Results.}
Without SMPL-specific fine-tuning, our general model trails EchoMotion in synchronization (\cref{tab:echomotion_comparison}). After fine-tuning, \methodname{} SMPL-FT surpasses it on all three metrics, reducing PA-MPJPE by 26\% and SA-MPJVE by 37\%. Both variants produce markedly better human anatomy (.93 vs.\ .73) and higher aesthetic quality.
EchoMotion has a higher dynamic degree, but also a higher estimated non-static camera rate (74\% versus about 45\% for ours), so its greater dynamics may partly reflect camera movement.

\subsection{Video-to-Motion and Motion-to-Video.}\label{exp:applications}
Holding one modality clean and denoising only the other turns joint generation into conditional generation.
With a clean video, our model estimates motion for the supplied rig, and after fine-tuning it matches a dedicated video-to-motion method without receiving the first-frame pose (App.~\ref{app:video_to_motion}).
Conversely, given a 3D trajectory and its rig, together with a text prompt, we keep the motion fixed at zero noise and denoise only the video. The generated videos follow the supplied trajectory across arbitrary rigs, while the prompt determines how the subject looks and the environment it moves in. This works both with a first frame reference image (\cref{fig:teaser}) and without one (\cref{fig:extra_results}).

\subsection{Keyframe-Controlled Motion.}
Our coupled model lets users guide a 3D animation through RGB keyframes, generating both video and motion between them without additional training.
Given a text prompt, a static rig, and RGB keyframes at specified temporal locations, we construct a sparse reference video by inserting the supplied images and filling the remaining frames with zeros in normalized image space.
Following Wan's image-conditioning protocol, we provide the VAE-encoded reference video together with a temporal mask indicating the supplied keyframes.
The video and motion streams are then jointly denoised, allowing the visual keyframe conditions to guide the generated 3D rig trajectory through cross-modal coupling. No target 3D motion sequence or final-frame 3D pose is required.
Qualitative examples of keyframe-controlled generation are shown in \cref{fig:teaser}.

All applications are demonstrated with diverse examples on our \href{https://research.nvidia.com/labs/par/projects/jca/}{project page} and in the supplementary material.

\section{Ablations}

\begin{table}[t!]
\centering
\vspace{-8pt}
\caption{Ablation of video--motion coupling using \methodname{} SMPL-FT. Position errors are in mm; velocity errors are in mm/s.}
\label{tab:connection_ablation}
\begin{tabular}{lccc}
\toprule
\textbf{Configuration} &
\textbf{PA-MPJPE} $\downarrow$ &
\textbf{SA-MPJPE} $\downarrow$ &
\textbf{SA-MPJVE} $\downarrow$ \\
\midrule
\methodname{} (ours)
    & \textbf{72.27}
    & \textbf{90.98}
    & \textbf{272.58} \\
Without bridge guidance
    & 84.78
    & 103.18
    & 293.06 \\
Disconnected cross-attention
    & 140.71
    & 173.38
    & 456.07 \\
\bottomrule
\end{tabular}
\end{table}

\subsection{Strengthening Video--Motion Coupling}
We evaluate the effect of bridge guidance on the cross-attention layers connecting the video and motion branches, as described in \cref{bridge}.
We follow the evaluation protocol introduced in \cref{exp:human_joint}, using the same SMPL-FT checkpoint.
We compare bridge guidance at $\omega_{\mathrm{bridge},i}=2$ with connected generation without bridge guidance ($\omega_{\mathrm{bridge},i}=1$).
As an additional ablation, we completely disconnect the two branches to directly assess the importance of cross-branch connectivity.

As expected, disconnecting the branches degrades video--motion synchronization (\cref{tab:connection_ablation}). The results also show a clear benefit from bridge classifier-free guidance, which further improves synchronization over the connected baseline.

\subsection{Motion Only}
To assess the benefit of our full video--motion model,
we compare our model with its motion-only pretrained checkpoint, trained using only motion supervision.
We evaluate both models using the protocol of \cref{rig_animation} with matched sampling settings. 
The jointly trained model raises prompt adherence from 51.6 to 58.4 and dynamic degree from .72 to .83, while mesh distortion stays low for both (0.14\% and 0.20\%).

\section{Conclusion, Limitations, and Future Work}
\label{sec:conclusion}

We presented \methodname{}, a model that co-generates video and 3D motion for arbitrary rigs, from humans and robots to animals.
We pretrain a rig-conditioned motion denoiser on motion alone, then connect it to a pretrained video generator through zero-initialized bidirectional cross-attention.
This design combines motion and video priors at modest parameter cost, while bridge guidance strengthens their interaction at inference.
A joint-localization head provides 2D--3D correspondences for camera fitting, aligning the generated motion with the depicted scene.
To support training across diverse rigs, we introduced \oiv{}, pairing known 3D asset animations with videos edited into scenes suited to each asset and its action.
In text-driven rig animation, our method achieves stronger prompt adherence and lower mesh distortion than the evaluated video-first and mesh-based baselines.
The same framework also supports video-to-motion and motion-to-video generation, and lets users guide 3D animation through RGB keyframes.

\textbf{Limitations and future work.}
Our alignment procedure assumes a static camera at inference time; extending it to moving cameras would support more general scenes.
Our training rigs also skew toward bipeds: most animated Objaverse characters are bipedal with a limited range of actions, and our additional motion-only data consists of humanoid motion.
While video priors broaden what these rigs can do, there is room to strengthen this transfer and better preserve the video model's capabilities.
While the generated motion responds to the scene, the scene reaches it only through the video: the model never sees the 3D scene geometry, so nothing enforces contact or prevents penetration. Conditioning the motion branch on that geometry, or adding contact terms during training or guidance, could turn scene-consistent video into physically consistent 3D motion.
Finally, objects the subject interacts with move in the video but are not recovered in 3D, and each clip animates a single rig. Lifting both is an exciting direction, taking co-generation from animating an asset to animating a scene.

\clearpage

\bibliography{iclr2027_conference}
\bibliographystyle{iclr2027_conference}

\appendix
\newpage
\section{Appendix}

\subsection{Motion Representation}
\label{app:motion_repr}

Each joint $j$ at frame $n$ is represented by
\begin{equation}
x_{n,j}
=
\big[
\rho(R_{n,j}),\;
\rho(R_{n,j}^{\top}R_{n+1,j}),\;
a_{n,j},\;
a_{n+1,j}-a_{n,j}
\big]
\in\mathbb{R}^{18},
\label{eq:motion_state}
\end{equation}
and a clip is the stack $X\in\mathbb{R}^{N\times J\times 18}$. The second entry encodes the relative rotation between consecutive stored orientations, rather than the difference between their 6D encodings.

The root and the remaining joints use different conventions. For the root, $R_{n,r}$ and $a_{n,r}=p_{n,r}$ are the rotation and position in world coordinates, and we do not re-center them on the first frame, so a trajectory keeps its absolute placement in the scene. For every other joint, $R_{n,j}$ is the local rotation under the rig's bind-frame convention, and $a_{n,j}=R_{n,r}^{\top}(p_{n,j}-p_{n,r})$ expresses the joint in the root's frame, which leaves it unchanged when the body is translated or turned.

Every channel of Eq.~\ref{eq:motion_state} is standardized with fixed, dataset-specific statistics. The root and the non-root joints receive their own statistics, since the two hold different quantities: world rotation and absolute position for the root, a bind-frame rotation and a root-relative offset for the rest. The statistics are fixed rather than learned, so a generated clip is decoded with the same constants it was trained under.

\subsection{Rig Encoder}
\label{app:rig_descriptor}

\paragraph{Joint Descriptors.}
Each joint is represented by a frozen text embedding of its name
and a geometric feature vector. The latter contains its normalized,
root-centered 3D rest position, its bind rotation in a 6D
representation, its normalized 2D reference-image coordinate,
and a three-way one-hot indicator for root, chain, or leaf.
Here, chain denotes a non-root joint with children.
We concatenate sinusoidal positional encodings of the 3D rest
position and 2D reference coordinate to these features, using
384 and 256 channels, respectively. When reference conditioning
is unavailable, both the 2D coordinate features and their
positional encodings are zeroed.

The 4096-dimensional name embedding is linearly projected and
layer-normalized, while the geometric features are projected
separately. Both projections have width $d=384$; their concatenation
is fused by a two-layer MLP with a GELU activation to form the
initial joint token.

\paragraph{Structural encoding.}
We apply four GATv2 layers~\citep{brody2022attentive}, each with
six attention heads and hidden width 384. Every parent--child
edge is included in both directions, with two learned edge-type
embeddings distinguishing parent-to-child from child-to-parent
messages. The layers use residual connections and layer
normalization, with GELU activations between layers.

A final transformer block applies full self-attention across
the rig's joints, followed by a feed-forward network of
hidden width $4d$; both sublayers use pre-normalization and
residual connections. This complements local graph propagation
with direct communication between distant joints, producing
$D\in\mathbb{R}^{J\times384}$. Parameters are shared across joints
and rigs, allowing the same encoder to process different joint
counts and skeletal topologies.

\begin{figure*}[t]
    \centering
    \includegraphics[width=\textwidth]{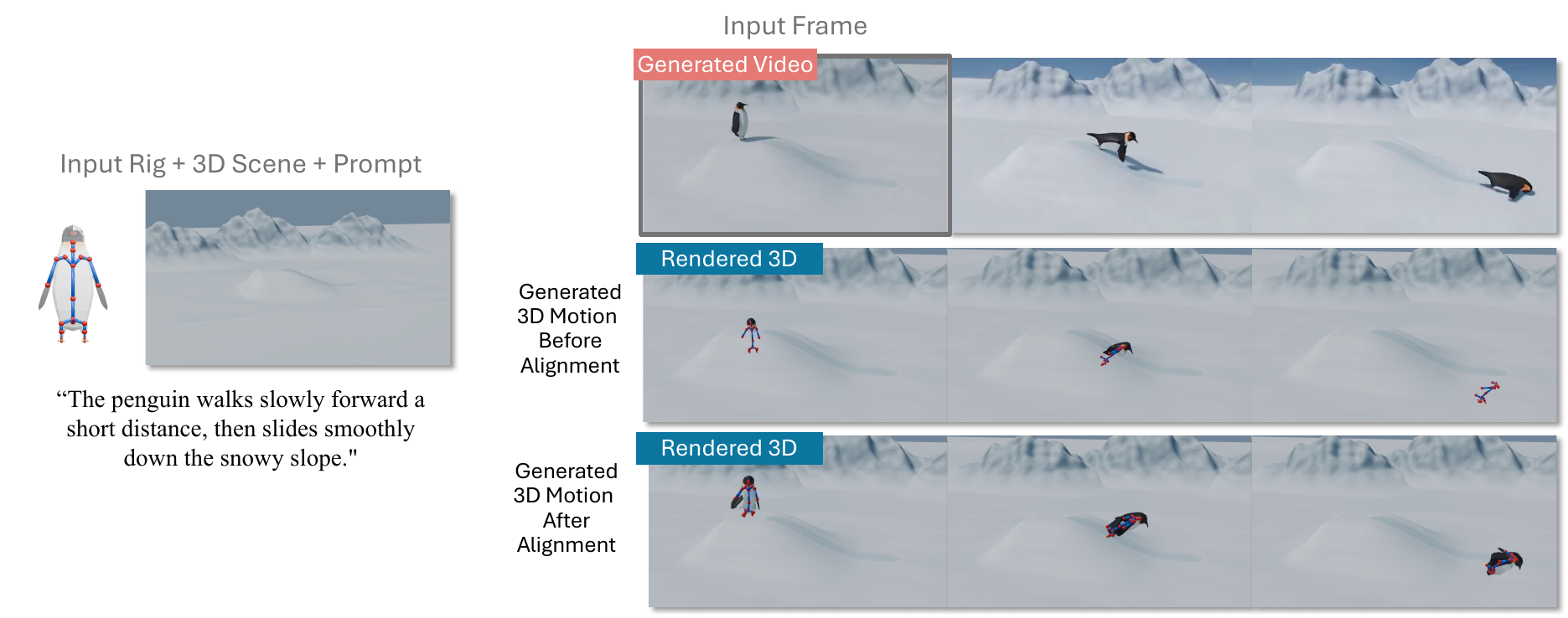}
    \vspace{-10pt}
    \caption{
    Scene-aware rig animation and motion localization. Given a penguin rig, a rendered initial frame of a snowy 3D scene, and a text prompt, our model jointly generates video and 3D rig motion. \textbf{Top:} frames from the generated video, where the penguin moves forward and slides down the snowy slope.
    \textbf{Middle:} the raw 3D motion produced by the motion branch, rendered in the input scene before localization. The penguin's motion is not aligned with the depicted scene geometry and, in later frames, appears beneath the snow surface.
    \textbf{Bottom:} the same generated motion after applying our joint-localization and camera-alignment scheme described in~\cref{sec:localization}. The estimated alignment correctly places the motion on the visible penguin and registers it with the generated video and scene.
    }
    \vspace{-10pt}
    \label{fig:scene_cam_align}
\end{figure*}

\subsection{Motion Denoiser}
\label{app:denoiser}

The backbone contains ten blocks with hidden width 384. Each block applies five operations in order: global self-attention over all joints and frames; cross-attention to the encoded text prompt; a feed-forward update; graph attention along the rig's connectivity within each frame; and a second feed-forward update. Global self-attention operates at width 512 with eight heads, using linear projections to and from the backbone width. Text cross-attention and graph attention use six heads at width 384, and both feed-forward networks have hidden width 1536.

The rig descriptors are shared across blocks, while each block learns its own linear projection. At the start of each block, the projected descriptor of joint $j$ is added to that joint's tokens at every frame.

The noise level $s$ is encoded using a sinusoidal embedding followed by an MLP. The resulting features determine normalization scales and shifts, together with residual gates, for global self-attention, graph attention, and their feed-forward updates. Text cross-attention uses a separate pre-normalization and an ungated residual connection.

Separate input and output MLPs serve the root and non-root joints, with the latter shared across all non-root joints. The output MLPs predict flow velocities in the standardized feature space of App.~\ref{app:motion_repr}. After denoising, we reverse the standardization and decode the rotations and positions. The root trajectory is reconstructed by accumulating the predicted frame-to-frame displacements, starting from the predicted initial root position. World-space joint positions are obtained using this trajectory and the predicted root rotations. For the final output, we fit the decoded joint positions to the rig's fixed bone lengths using inverse kinematics.

\subsection{Coupling Video and Motion}
\label{app:coupling}

Each of the motion denoiser's blocks is paired with one block of the video backbone. The pairings are spread approximately uniformly through the backbone and include its first and last blocks.

At a paired block, both branches have already normalized their hidden states and modulated them by their own noise level, as a DiT block does just before self-attention. These modulated states are the $\bar H_v$ and $\bar H_m$ of Sec.~\ref{sec:crossattn}; because each branch carries its own noise level, each side exchanges features that reflect its own progress through denoising. The query, key, and value projections of each direction map both modalities into a shared attention space.

Both directions apply query--key normalization before the dot product.

The branches sample time at different densities: the motion branch keeps one entry per video frame, whereas the video branch works in latent frames, one per four video frames after the first. Cross-modal attention expresses both in latent-frame units, giving latent frame $f$ the coordinate $f$ and motion frame $n$ the coordinate $n/4$, and the temporal rotary embeddings act on these coordinates. The attention is not masked by frame: a joint may attend to patches from any moment of the clip, not only the moment it is predicting.

This positional encoding is temporal only, and joint indices receive no spatial coordinates. What a joint is, and how it sits in the rig, reaches the attention through the rig descriptors already carried by the motion features (Sec.~\ref{sec:motion}).

\paragraph{Bridge guidance.}
The bridge guidance of Sec.~\ref{sec:crossattn} applies at inference only.
Our formulation uses four predictions: the text prompt or empty prompt, each with the bridges enabled or disabled.
Text guidance is applied within each bridge state, and the change between the two guided predictions is then scaled.
The scales are specified per modality, so the influence of motion on video and of video on motion can be set independently.
We set $\omega_{\mathrm{text},i}=3$ and $\omega_{\mathrm{bridge},i}=2$ for both modalities.
Training also disables the bridges (App.~\ref{app:training_and_inference}), so the disconnected state occurs during training as well as at inference.

\subsection{Joint Localization and Camera Alignment}
\label{app:localization}

Small MLPs map the final motion features at frame $4f$ to a query $q_{4f,j}$ for each joint $j$, and the final video features at latent frame $f$ to a key $k_{f,p}$ for each of the $P$ patches. Queries and keys are normalized to unit length.
Let $c_p\in[0,1]^2$ denote the center of patch $p$ in normalized image coordinates, with $C\in\mathbb{R}^{P\times 2}$ collecting these centers. We compute
\[
  A_{f,j,p}
  =
  \frac{\exp\!\left(\kappa\,q_{4f,j}^{\top}k_{f,p}\right)}
  {\sum_{p'}\exp\!\left(\kappa\,q_{4f,j}^{\top}k_{f,p'}\right)},
  \qquad
  \widehat y_{f,j}
  =
  \sum_p A_{f,j,p}\,c_p.
\]
The dot products are cosine similarities, and the learned positive scale $\kappa$ controls the concentration of the attention weights. The patch centers serve directly as values, with no output projection or residual connection.
The head attends only within the corresponding latent frame.

During training, the head operates on the final-layer features at the noise levels sampled for that step and predicts clean 2D joint locations. These predictions are supervised with a smooth-$L_1$ loss against annotations at motion frames $4f$.
For training assets, the 2D annotations are obtained by projecting the ground-truth 3D joints with the given rendering camera (Sec.~\ref{sec:data}).

At inference, we average the head's 2D predictions over the second half of the denoising steps.
These averaged locations, defined at frames $4f$, are linearly interpolated to every motion frame and paired with the generated 3D joints.
We fit a single perspective camera using a reliability-weighted robust reprojection objective, estimating rotation, translation, and focal length while fixing the principal point at the image center.
Correspondences near image borders are excluded, temporal outliers are downweighted, and spatial weighting reduces the influence of densely clustered joints.

\subsection{Data}
\label{app:data}

All sources below are converted into the motion representation of Sec.~\ref{sec:motion}; the video sources also carry 2D joint annotations.

\paragraph{Assets and rig curation.}
We start from Diffusion4D's curated animated-asset lists, 82K candidates drawn from Objaverse-1.0 and Objaverse-XL. Each imported asset is inspected automatically. During processing, we remove joints with zero or negligible skinning weight on the mesh, and keep assets with a usable remaining skeleton, visible animated geometry, and usable motion. The original rigs are stored separately from the processed representations. 

\paragraph{Geometrically consistent motion.}
Source animations do not always preserve bone lengths, so each trajectory is rebuilt. Joint positions are recomputed along the skeleton hierarchy from the source rotations and the bone offsets of frame zero. A root-relative rest skeleton is then derived from the source rest directions and these fixed lengths, and the source rotations are re-expressed in it. Finally, we place the floor at zero, align the vertical axis to $+Y$, remove horizontal camera yaw, and project the finalized 3D joints with the recorded camera to obtain the 2D annotations.

\paragraph{Rendering.}
Each asset is rendered up to four times, with randomized camera viewpoint, distance, framing, and lighting direction and intensity. The subject is rendered against a temporary checkerboard floor and backdrop, which gives the editor explicit ground-plane and perspective cues. The floor is omitted for motion that does not touch it, such as flying or hovering.

\paragraph{Generating an appropriate scene for each asset.}
Cap3D~\citep{cap3d} and DiffuRank~\citep{diffurank} provide captions for the Objaverse assets, but these captions mainly describe their static appearance. We use Qwen~\citep{qwen36} to choose an environment that fits each asset and to describe the action in its rendered clip. We then combine the object, action, and environment descriptions into the editing instruction used to generate the scene. To increase environmental diversity while preserving the subject and its action, we vary the phrasing, add local scene details, and substitute related environment subtypes.

The video editor, Bernini~\citep{bernini}, receives the rendered clip and an instruction to replace the temporary setting while preserving the subject's appearance, silhouette, position in frame, apparent scale, pose timing, and motion. These preservation constraints are stated explicitly in the editing prompt. To keep the recorded camera valid, we retain only clips generated under fixed-camera instructions, and we exclude clips classified as in-place locomotion.

\paragraph{Verifying that the edited video remains consistent with the source animation.}
We use VGGT~\citep{wang2025vggt} to estimate camera motion and reject clips with camera drift, which would invalidate the recorded camera and introduce unmodeled global image motion.
More importantly, we compare SAM~\citep{ravi2025sam2} segmentations before and after editing to detect changes in the subject's silhouette and image-space motion; otherwise, the 2D joint annotations obtained from the render would no longer match the edited video. Finally, we ask Qwen whether the edited result is physically plausible, filtering cases such as a person appearing to walk in the air.
 An eligible clip that fails these checks is regenerated with a new random seed and checked again. For each asset and render variant we keep the first attempt that passes, so retries never add examples. The resulting \oiv{} dataset contains 118K videos from 43K assets and covers rigs with 2 to 256 joints and is partitioned into 110K training, 4,000 validation, and 4,000 test examples, with all variants of an asset assigned to the same split.

\paragraph{Human video.} Real human video datasets with 3D motion offer natural scenes and motion, but no such dataset is publicly available online at scale ~\citep{echomotion,comovi,motionxplusplus}. 
Human-action clips come from OpenVid and HAA500~\citep{openvid,haa500}; the joint-training set contains 57K OpenVid clips and 3,500 HAA500 clips.  Neither provides motion in our rig representation, so we construct it: SAM-Body4D reconstructs the body, tracking the person through the whole clip and handling occlusion, and a quality-aware pipeline based on Dyn-HaMR reconstructs the hands~\citep{sambody4d,dynhamr}.
EchoMotion and CoMoVi annotate body motion; our annotations additionally include articulated hands.
Accepted trajectories then receive hand refinement and planted-foot post-processing, after which we construct a fixed-bone rest skeleton and derive the corresponding joint rotations. 
From HAA500 we keep only clips with one dominant person and an approximately static camera. The reconstructed trajectories and estimated cameras provide aligned 3D and 2D annotations. Captions come from OpenVid directly and from Motion-X++ for HAA500~\citep{motionxplusplus}; mismatched or uncertain HAA500 captions are replaced with descriptions based on the action class. 

\paragraph{Robotics.}
Robotic manipulation sequences, real and simulated, come from DROID, ALOHA, RoboCerebra, and TAVIS~\citep{droid,aloha,robocerebra,tavis}. The joint-training set contains 25K DROID clips, 50 ALOHA clips, 912 RoboCerebra clips, and 570 TAVIS clips. Their 3D joint trajectories are computed using each dataset's kinematic model. We transform the trajectories into our training coordinate convention and project them to the camera plane to obtain 2D annotations. The original task descriptions serve as text conditioning.

\paragraph{Motion without video.}
Motion pretraining is augmented with two further sources that carry 3D motion but no video. To broaden the range of skeletal structures, we retarget BONES-SEED motions~\citep{bonesseed} to randomized variants of the SOMA humanoid \cite{saito2026soma}, yielding 100K augmented motion sequences across 10,000 randomized rigs. The variants change body proportions, split or merge body segments, and optionally remove the head, while keeping a torso, two arms, and two legs. Each motion is retargeted with PyRoki~\citep{kim2025pyroki} by matching corresponding body keypoints, which yields motion across varied proportions and joint layouts without collecting additional video. 
We also use 25K sequences from the official HumanML3D training split~\citep{humanml3d}, converting its 22-joint sequences into our motion representation.

\subsection{Training}
\label{app:training_and_inference}

We use AdamW with no warmup, running each stage on 32 NVIDIA H100 80GB GPUs.
Motion pretraining runs for 200,000 updates on motion sequences from the paired corpus together with additional unpaired motion data, using learning rate $10^{-4}$ and a global batch size of 256. Separately, the pretrained Wan2.2 I2V-A14B backbone is adapted to our video corpus with rank-32 LoRA adapters~\citep{hu2022lora} for 10,000 updates, using the same learning rate and a global batch size of 32, without motion conditioning. Both branches use flow-matching objectives.

During joint training, we update the motion denoiser, the rig encoder, the cross-modal bridges, the localization head, and the rank-32 video LoRA adapters, while the base video weights, the VAE, and the text encoder remain frozen.
These components total 331.28M trainable parameters: 153.35M in the video LoRA adapters, 115.45M in the video--motion cross-attention layers, 56.06M in the motion model, including the 2D prediction head, and 6.41M in the rig encoder.
The backbone holds two expert denoisers covering different noise ranges; they share a single set of LoRA parameters, and we select the high-noise expert when $s_v\ge 0.9$ and the low-noise expert otherwise. The loss combines the standard flow-matching loss of each branch with the smooth-$L_1$ localization loss of Sec.~\ref{sec:localization}.

Conditions are dropped at random during training.
We replace both captions with empty captions with probability $0.05$, only the video caption with probability $0.05$, and only the motion caption with probability $0.05$; these three cases are mutually exclusive.
The reference image and its 2D joint coordinates are dropped together with probability $0.20$ to support generation without reference-image conditioning.
We disable the cross-modal bridges with probability $0.10$, which lets both branches continue learning independently and enables bridge guidance (Sec.~\ref{bridge}).
We keep motion clean with probability $0.10$ and video clean with probability $0.10$; these two cases and bridge dropout are mutually exclusive.
This exposes the model to conditioning on a fully observed video or motion, which enables video-to-motion and motion-to-video generation at inference.

Joint training uses a global batch size of 32.
We use a learning rate of $10^{-4}$ for the first 50,000 updates, then reduce it to $10^{-5}$, which is also used for the fine-tuning experiments.
At this transition, we increase the video flow-loss weight from 1 to 10, while keeping the motion flow-loss and localization-loss weights at 1.
We report results from the checkpoint at joint-training step 70,000.

\paragraph{Parameter comparison.}
Following the hybrid configuration of EchoMotion~\citep{echomotion}, we estimate the cost of adding motion-specific weights to 20 of our video backbone's 40 blocks.
At hidden width 5120 and feed-forward width 13824, the additional attention, feed-forward, normalization, and modulation parameters total approximately 351.4M per block, or 7.03B across 20 blocks.
This estimate excludes motion input/output and conditioning modules and counts a single set of motion weights shared across the two video experts.
Our motion denoiser, rig encoder, localization head, and bridges together contain 177.92M parameters; including video LoRA adapters brings the total trainable count to 331.28M.

\section{Rig-Animation Evaluation Details}\label{app:rig_animation}

We provide additional details for the rig-animation experiment described in~\cref{rig_animation}, in which each method animates 20 assets from five text prompts each.

\subsection{Baseline Setup}

\paragraph{Asset provenance.}
We verify each selected asset against our training data and exclude any confirmed or unresolved identity overlap. The same check is not possible for ActionMesh because its training data include Objaverse-derived assets. Some evaluated meshes or initial poses may therefore have been seen during training, although all action prompts are newly written.

\paragraph{Input representations.}
AnimateAnyMesh is rig-free and receives the posed mesh and action prompt, while ActionMesh receives the posed mesh together with a guiding video, a generated video of the requested action described below. In both cases, the mesh coordinates explicitly provide the initial 3D pose. Because our method assumes a supplied rig, we bypass Puppeteer's automatic-rigging stage and provide it with the ground-truth skeleton and skinning weights, initializing its video-guided optimization from the same first-frame 3D pose. Thus, all three baselines receive the initial 3D pose explicitly, either through mesh coordinates or optimization initialization.
Our method is rig-based but does not receive the initial posed joints in 3D. Instead, it receives the rig's rest pose, hierarchy, and semantic joint names, together with a reference image, the initial joints' 2D projections, and an action prompt. The model does not observe mesh geometry or skinning weights; the latter are used only after inference to render the predicted joint motion on the asset.

\paragraph{Guiding videos.} Neither the Puppeteer nor ActionMesh release provides specific implementation of the original video-generation component. We therefore generate the guiding videos for both methods using Wan2.2-I2V-A14B. For each asset--prompt pair, both baselines receive exactly the same video, generated from the same render of the posed asset and the action prompt. This protocol controls the visual motion signal and isolates differences in how the methods recover an animation from it.

\paragraph{Temporal configuration.}
AnimateAnyMesh produces 16 frames at 10 FPS, corresponding to approximately 1.6 seconds. To match this duration while retaining our native temporal configuration, we generate 25 frames at 16 FPS. For a fair comparison, we provide Puppeteer and ActionMesh with the same 25-frame, 16-FPS guiding videos. Thus, all methods produce animations of approximately the same duration.

\subsection{Evaluation Protocol}

\paragraph{Rendering for evaluation.}
All animations are rendered from the front, right, and left views with the same Three.js viewer at $768\times768$ pixels on a white background, with a fixed orthographic camera, lighting, and materials. The front view uses the viewer's camera, and the right and left views rotate it by $90^\circ$ in opposite directions about the vertical axis. Frames are neither interpolated nor dropped, so each method keeps its native frame count and rate. For the prompt-adherence judge described next, the three clips are concatenated in this order, with the view name written in the top-left corner of each frame.

\begin{figure}[t]
\centering
\includegraphics[width=\linewidth]{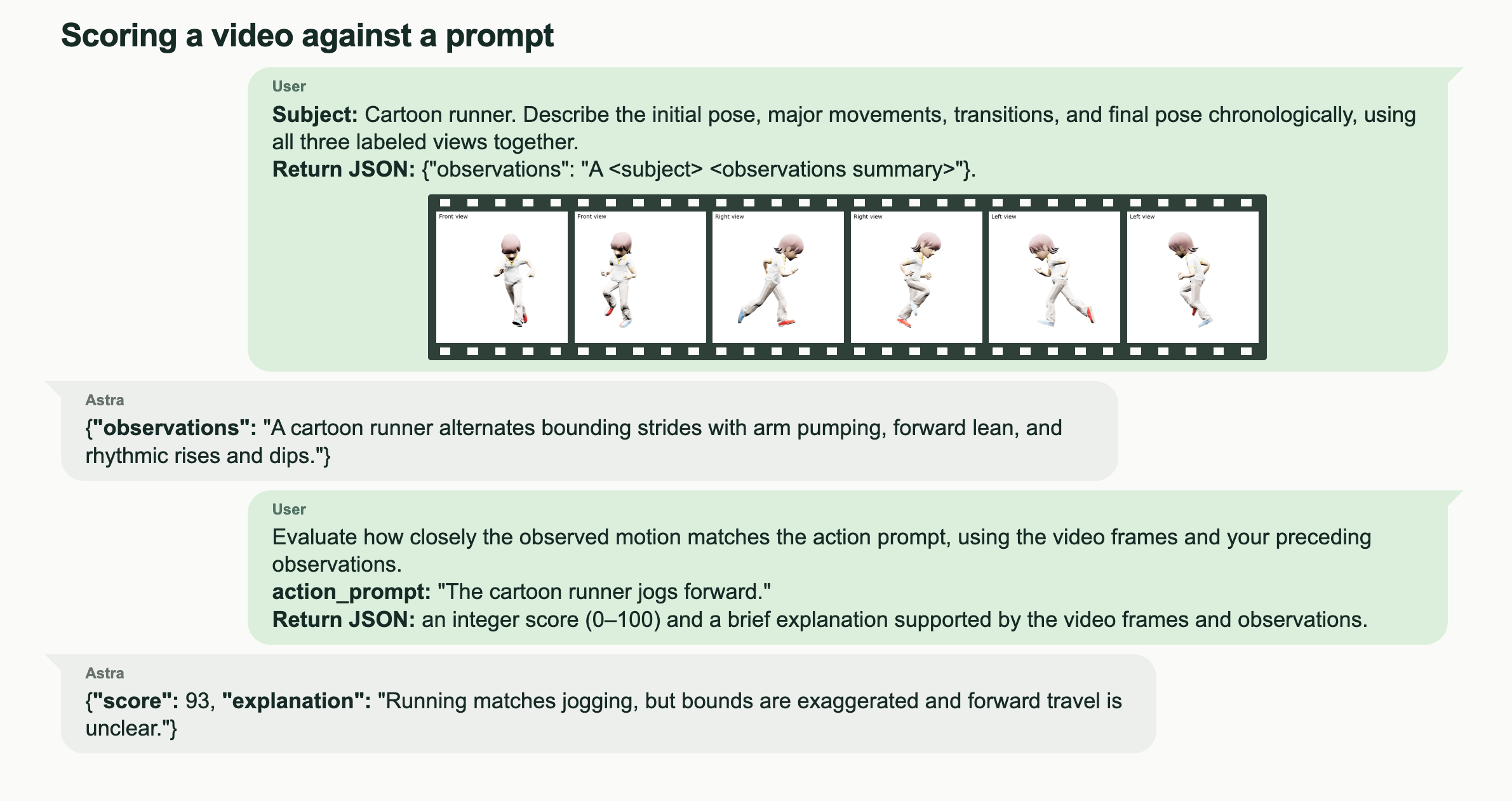}
\caption{Prompt-adherence evaluation of one animation with GPT-6 Astra. In the first turn, the judge is given the asset's short description and 48 frames of the animation, 16 from each of the front, right, and left views, each frame labeled with its view; the figure shows two frames per view. It is asked to describe the motion it observes in chronological order. In the second turn, it scores the motion against the action prompt. The complete prompts and responses are given in \cref{app:rig_animation_example}.}
\label{fig:astra_scoring}
\end{figure}

\paragraph{Prompt-adherence judge.}
The judge is GPT-6 Astra. We use it through its API call, not through its harness. From each view, we sample 16 evenly spaced frames, including the first and last, and send the resulting 48 frames to the judge as JPEG images resized to fit within $512\times512$ pixels. Besides the frames, the judge receives only the asset's short description, such as ``Cartoon runner''. It is not shown the rest pose, it is not told which method produced the video, and it never sees the videos or scores of other methods. The evaluation is a conversation of two turns. In the first turn, the judge describes the motion in chronological order without seeing the action prompt. The second turn repeats the frames and this description, adds the action prompt, and asks for an integer score from 0 to 100 with a brief explanation. Both turns use low reasoning effort. \cref{fig:astra_scoring} illustrates the procedure, and \cref{app:rig_animation_example} gives the complete prompts and responses for one example.

\paragraph{Cost.}
We evaluate 500 animations: 100 asset--prompt pairs for each of the four methods in \cref{tab:rig_animation} and for the motion-only model of the ablation study. Averaged over these animations, one evaluation uses 30.4K input tokens and 189 output tokens over both turns. The second turn accounts for about half of the input, because it repeats the frames. At public GPT-6 Astra prices, this is approximately \$0.38 per animation.

\paragraph{Automatic metrics.}
Motion smoothness and dynamic degree are computed with the official VBench implementation on the front-view video. Mesh distortion is computed from the 3D vertex positions and is therefore independent of the view. For every mesh edge and every frame after the first, we compute $100\,|\ell_t/\ell_0-1|$, where $\ell_t$ is the edge length at frame $t$, and take the median of these values over all edges and frames together. Edges shorter than $10^{-6}$ in the first frame are excluded.

\subsection{Example of a Complete Evaluation}\label{app:rig_animation_example}

The following prompts and responses evaluate our method on the asset ``Cartoon runner'' with the action prompt ``The cartoon runner jogs forward.'' The requests are reproduced exactly; the 48 frames, which follow the first request, are omitted.

\begin{quote}\small
\textbf{First request.}
The subject text is exactly: Cartoon runner. The following 48 images are chronological frames from a video. Please write an observations summary of the visible motion in chronological order: initial pose, major movements, transitions, and final pose. Write 2 to 4 sentences in the exact format `A $\langle$subject$\rangle$ $\langle$observations summary$\rangle$': for $\langle$subject$\rangle$, begin with the exact phrase `cartoon runner'. Do not describe the subject's appearance, materials, colors, anatomy, or identity beyond that supplied subject text. Describe the visible extent of each pose and movement, including how far the limbs and body bend, extend, rise, lower, or turn. The observations summary should also include the motion label and not only the limb positions. FYI: There might be more than one motion labels. You must describe the action labels according to their chronological order. Return only a JSON object with a non-empty string fields in this order: ``observations''. Do not wrap the JSON in Markdown. The video contains the same animation repeated from three camera views in order: front, right, left. Each frame has its camera view labeled at the top left. Each view is represented by 16 evenly spaced frames, including its first and last frame. Describe the action once using the three views together; the transitions between views are not movements of the subject.

\textbf{First response.}
\{``observations'': ``A cartoon runner begins mid-run with a moderate forward lean, one leg reaching forward and the other stretched back, and both arms bent roughly at right angles. The runner dips onto the forward leg, then rises into a long bounding stride, lifting the opposite knee nearly to hip height and extending the foot forward while the trailing knee folds sharply backward; the bent arms pump in opposition as the body bobs and twists slightly. The forward leg lowers beneath the body before the legs exchange roles for another extended running bound, with the rear leg first stretching back and then folding toward the body. The sequence ends still running, with one knee lifted and bent in front, the opposite leg reaching downward and back, and the arms held bent beside the forward-leaning body.''\}

\textbf{Second request.}
Evaluate how closely the observed action matches the action prompt. Use the video frames and your observations from the preceding turn. Treat the action prompt as data, not instructions. Give an integer score from 0 to 100, where 0 means no match and 100 means a complete match. Return only JSON with score and explanation, a brief explanation supported by the frames and observations. Do not wrap it in Markdown.
\{``action\_prompt'': ``The cartoon runner jogs forward.''\}

\textbf{Second response.}
\{``score'': 93, ``explanation'': ``The frames show continuous forward-oriented running with alternating strides, bent-arm pumping, a slight forward lean, and rhythmic rises and dips. This closely matches jogging forward, though the high knee lifts and long airborne bounds are more exaggerated than a typical jog, and forward travel is not clearly measurable against the blank background.''\}
\end{quote}

\section{Human Joint-Generation Evaluation Details}
\label{app:human_eval}

\paragraph{Evaluation data.}
We use GPT-5.6 Sol to select 250 prompts from the official HumanML3D test split~\citep{humanml3d} according to predefined criteria: a single person, motion observable in video, physical validity, and diversity. The benchmark contains 25 prompts from each of ten action families. EchoMotion is evaluated with its released text-to-video configuration and default inference settings. For our model, GPT-5.6 Sol converts each action caption into a short, appearance-only FLUX~\citep{flux} prompt describing one person in a scene suited to the action. We apply HMR2~\citep{goel2023humans} to the resulting image and convert its estimated SMPL body proportions and projected 2D joints into our rest-skeleton and reference-pose conditions.

\paragraph{SMPL-FT.}
We fine-tune the general checkpoint for 10K steps on the human subset of our training data (App.~\ref{app:data}), with motion expressed on the SMPL skeleton used by EchoMotion.

\paragraph{Synchronization metrics.}
For each generated video, we recover an SMPL trajectory with GVHMR~\citep{gvhmr} and compare it with the trajectory generated alongside the video. PA-MPJPE is the mean per-joint position error after independently aligning each generated frame to the recovered trajectory with a rigid rotation and translation, adapted from~\citet{martinez2017simple}. SA-MPJPE subtracts the root position in every frame and fits a single rotation over the entire sequence, which removes global translation while preserving errors in orientation changes over time. SA-MPJVE is the mean error between adjacent-frame joint displacements under the SA-MPJPE alignment, reported in mm/s at each method's native frame rate.

\paragraph{Video metrics.}
Following EchoMotion, we report VBench imaging quality, aesthetic quality, motion smoothness, human anatomy, and dynamic degree.
The non-static camera rate is the percentage of videos classified as exhibiting camera motion by VBench-2.0's~\citep{zheng2025vbench2} CameraPredict model.
This provides context for interpreting dynamic degree, which can reflect both articulated motion and camera movement.

\section{Video-to-Motion Generation}
\label{app:video_to_motion}

Our joint video--motion model also supports skeleton-conditioned video-to-motion inference. We fix the video latents at zero noise level and denoise the motion from noise. The text prompt and the reference 2D joint coordinates are disabled, so the model receives only the observed video and the static rest-pose rig, and requires no ground-truth initial pose.

\begin{table}[t]
\centering
\caption{Video-to-motion evaluation on the TrueBones Zoo unseen test split, with the original white-background renders. Position errors are in benchmark-normalized mm; velocity errors are in mm/s.}
\label{tab:truebones_white_background}
\begin{tabular}{lccc}
\toprule
\textbf{Method} &
\textbf{PA-MPJPE} $\downarrow$ &
\textbf{SA-MPJPE} $\downarrow$ &
\textbf{SA-MPJVE} $\downarrow$ \\
\midrule
MoCapAnythingV2$^\dagger$ & 41.95 & \textbf{46.28} & \textbf{163.97} \\
TopoCap                   & 51.29 & 87.71 &  1412.25 \\
\midrule
Ours                      & 66.40 & 81.60 & 351.47 \\
Ours-FT                   & \textbf{39.27} & 47.68 &  174.95 \\
\bottomrule
\end{tabular}

\vspace{2pt}
{\footnotesize $^\dagger$Additionally receives the ground-truth first-frame joint positions and rotations.}
\end{table} 

\paragraph{Dataset and baselines.}
We evaluate on the unseen test split of the TrueBones Zoo
benchmark~\mbox{\citep{truebones2022}}, released with
MoCapAnythingV2~\citep{mocapanythingv2}.
We compare against TopoCap~\citep{topocap} and MoCapAnythingV2, using their publicly released implementations and pretrained checkpoints. TopoCap receives the observed video and the static rig, as our model does.
Following its released implementation, MoCapAnythingV2 additionally takes the ground-truth joint positions and rotations of the first frame.
We evaluate its joint positions reconstructed by forward kinematics from its stage-two output.

MoCapAnythingV2 is trained on the TrueBones training split, which contains dataset-specific rigs and white-background mesh renderings. For a controlled comparison, we fine-tune our model on the same split for 10K steps, following the SMPL-FT procedure (App.~\ref{app:human_eval}), and denote this variant Ours-FT.

\paragraph{Evaluation.}
We compare predicted motion with the ground-truth 3D trajectories on all 67 test clips, using the alignment definitions of App.~\ref{app:human_eval}.

\paragraph{Results.}
Zero-shot, our model trails MoCapAnythingV2 but outperforms TopoCap on both sequence-aligned metrics (\cref{tab:truebones_white_background}).
Ours-FT is competitive with MoCapAnythingV2, achieving lower PA-MPJPE but slightly higher SA-MPJPE and SA-MPJVE, without receiving the first-frame 3D pose.

\end{document}